\documentclass[letterpaper, 10 pt, conference]{ieeeconf}  

\IEEEoverridecommandlockouts                            

\usepackage{color}
\usepackage{graphicx} 
\usepackage[american]{babel}
\usepackage{subfigure}		
\usepackage{balance}
\usepackage{cite}
\usepackage{xcolor}
\usepackage{hyperref}
\usepackage{amsmath}
\usepackage{amssymb}

\usepackage{multirow}

\title{\LARGE \bf
CineMPC: Controlling Camera Intrinsics and Extrinsics for Autonomous Cinematography
}

\author{\centering Pablo Pueyo, Eduardo Montijano, Ana C. Murillo and Mac Schwager
\thanks{This work was supported by a DGA scholarship; Spanish projects PGC2018-098817-A-I00 and PGC2018-098719-B-I00 (MCIU/AEI/FEDER, UE), DGA T04-FSE; NSF grants CNS-1330008 and IIS-1646921; ONR grant N00014-18-1-2830, and ONRG-NICOP-grant N62909-19-1-2027.}
\thanks{P. Pueyo, E. Montijano and A. C. Murillo are associated with the Instituto de Investigaci\'on en Ingenier\'ia de Arag\'on, Universidad de Zaragoza, Spain 
\texttt{\small \{ppueyor, emonti, acm\}@unizar.es}}
\thanks{M. Schwager is associated with Dept. of Aeronautics and Astronautics, Stanford University, USA
\texttt{\small \{schwager\}@stanford.edu}}
}

\begin{document}

\maketitle
\thispagestyle{empty}
\pagestyle{empty}

\begin{abstract}
We present CineMPC, an algorithm to autonomously control a UAV-borne video camera in a nonlinear Model Predicted Control (MPC) loop.  CineMPC controls both the position and orientation of the camera---the camera extrinsics---as well as the lens focal length, focal distance, and aperture---the camera intrinsics.  While some existing solutions autonomously control the position and orientation of the camera, no existing solutions also control the intrinsic parameters, which are essential tools for rich cinematographic expression.  The intrinsic parameters control the parts of the scene that are focused or blurred, the viewers' perception of depth in the scene and the position of the targets in the image. 
CineMPC closes the loop from camera images to UAV trajectory and lens parameters in order to follow the desired relative trajectory and image composition as the targets move through the scene.

Experiments using a photo-realistic environment demonstrate the capabilities of the proposed control framework to successfully achieve
a full array of cinematographic effects not possible without full camera control.
\end{abstract}

\section{Introduction}
\label{sec_intro}
Unmanned Aerial Vehicles (UAVs) 
are becoming powerful platforms with high potential for applications far more sophisticated than aerial monitoring or personal entertainment. 
They are often equipped with high-quality cameras that allow them to record scenes from viewpoints that are challenging to capture using conventional recording devices. This fact has attracted the interest of the cinematographic industry, where more and more movies contain scenes recorded from flying vehicles. 
 To provide more sophisticated features to film-makers, it is essential 
 to make drones easier to maneuver and with automatic behaviours that are easy to configure by non-expert users. 
 \begin{figure}[!ht]
\centering
\begin{tabular}{cc}
    \includegraphics[width=0.46\columnwidth,height=2.4cm]{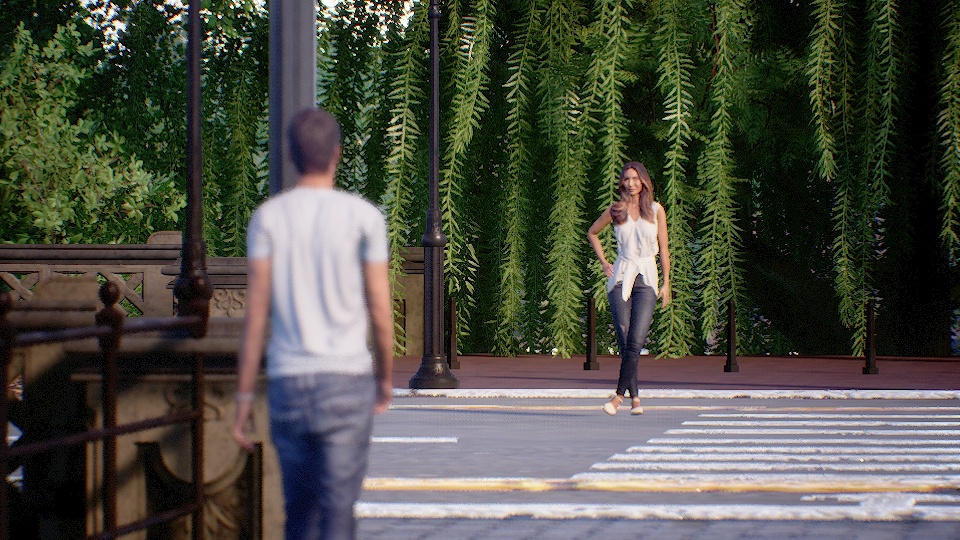} 
    &
     \includegraphics[width=0.46\columnwidth,height=2.4cm]{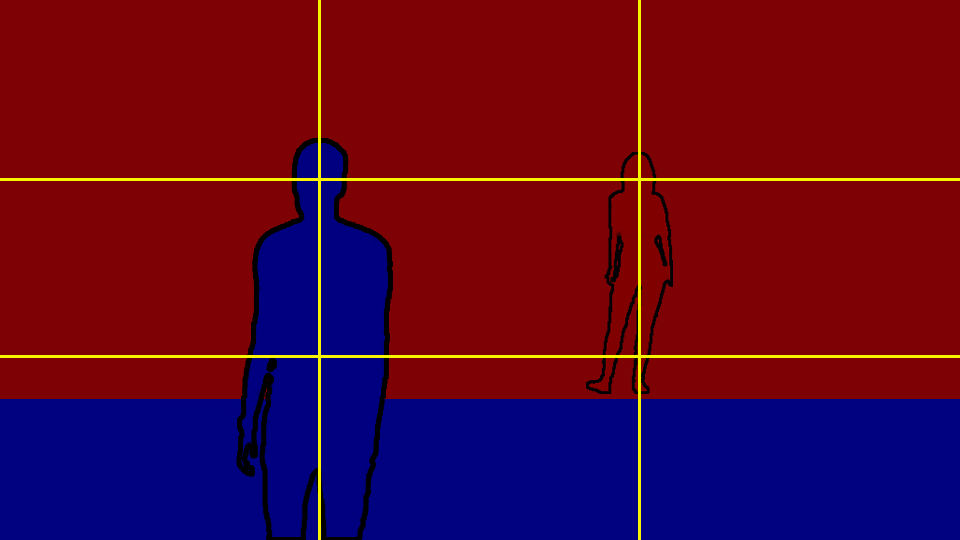}
 
    \\
    \footnotesize (a) CineMPC & \footnotesize (c) Instructions\\
       \includegraphics[width=0.46\columnwidth,height=2.4cm]{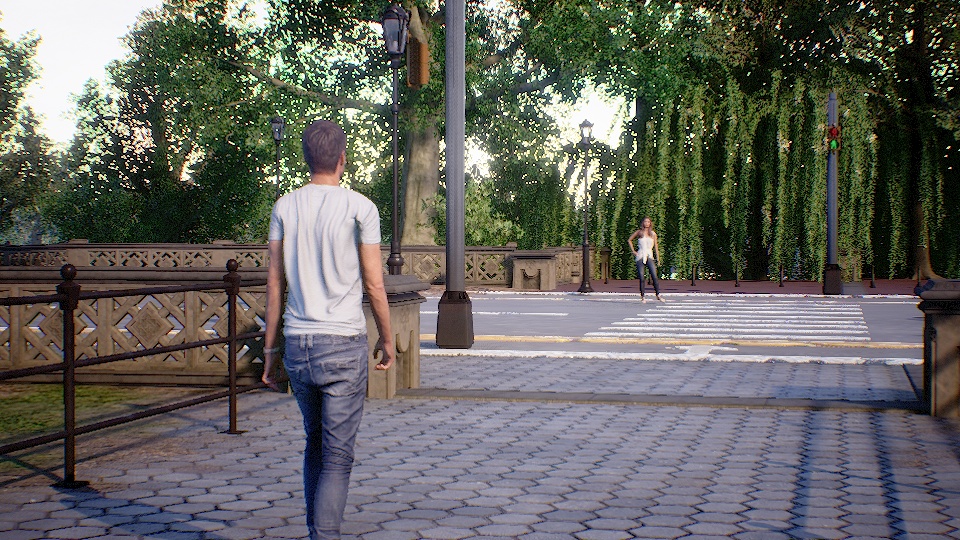}
    &
    \includegraphics[width=0.46\columnwidth,height=2.4cm]{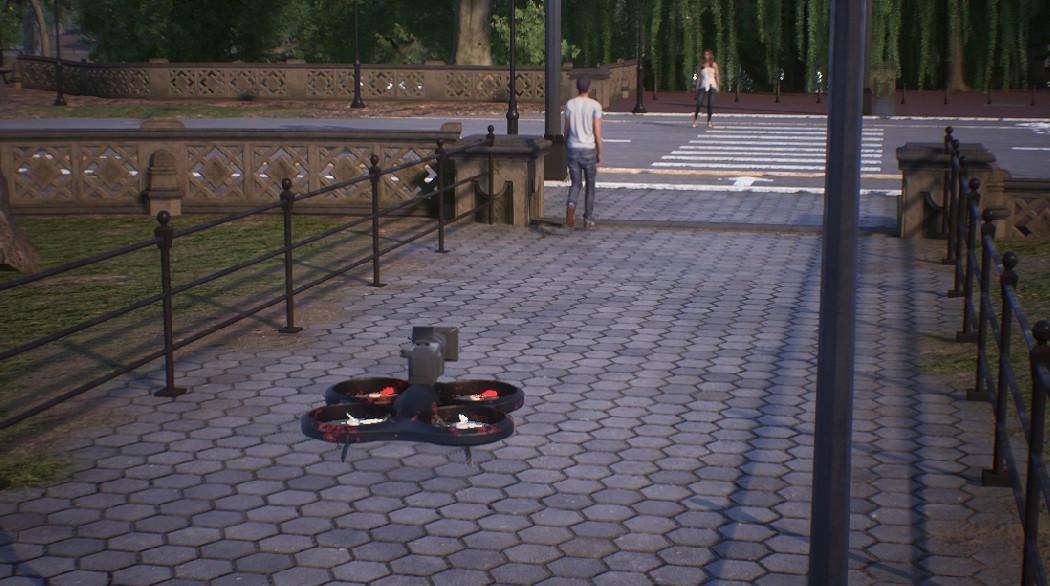}\\
    \footnotesize (b) Baseline & \footnotesize (d) Third person view\\
\end{tabular}
\caption{CineMPC in action. This example shows one frame of a recording. (a) First person-view of the drone when CineMPC controls the intrinsics of the camera (blurred foreground, sharper background and zoom helping to place the person in the image). (b) First person-view when the control problem does not include the intrinsics of the camera (all the scene homogeneously focused and not satisfying desired position in image). (c) Configuration input by the user to achieve this effect (red is focused area, blue is out of focus and yellow lines depict the desired image position of the top and lower parts of the target). (d) Third person-view filming drone including the elements of the scene. See supplementary materials for complete video demonstration.
}
\label{fig:main}
\end{figure}

 Existing literature offers several solutions for controlling drones autonomously to capture cinematic footage. However, to the best of our knowledge, existing methodologies only focus on the control of the extrinsic parameters of the drone and camera, namely, their position and orientation.
 This is because they rely on the pin-hole camera model in their control solutions, neglecting some of the most important cinematographic characteristics, e.g., depth of field, focal distance, and zoom. 
Cinematographic cameras allow real time modification of several intrinsic parameters to make these effects possible, and can be described by the thin-lens camera model \cite{lu2020camera}. CineMPC incorporates this camera model in the control problem, 
enabling it to automatically control the aforementioned cinematographic objectives together with the desired extrinsic configuration.

The core idea in this work is to adapt the classic cinematographic concepts~\cite{thompson2009grammar} to mathematical expressions that can be optimized using control techniques. 
Our main contribution is to control the drone position and orientation together with the intrinsic parameters of the camera lens in one unified control problem. 
Thanks to CineMPC, cinematographers can achieve a wider variety of effects and  configurations by specifying some artistic and composition guidelines (as shown in the example in Fig.~\ref{fig:main}). These specifications are optimized thanks to a Model Predictive Control (MPC) formulation that transforms them to instructions to autonomously control the drone and the camera while recording footage.

We show the potential of this approach with a photorealistic experiment using CinemAirSim \cite{pueyo2020cinemairsim}, an extension for cinematographic purposes of the photorealistic robotics simulator AirSim\cite{shah2018airsim}. The cameras embedded on the drones of CinemAirSim follow the thin-lens camera model. The experiments demonstrate the effect of applying the proposed control to enable a variety of changes in the depth of field, zoom, and framing recording strategies automatically.



\section{Related work}
\label{sec_related}
Several works focus on making the trajectory of the drone smoother while recording. Given a set of way-points, different MPC formulations are used to control the drone to avoid unstable trajectories while passing through the  points \cite{rousseau2018quadcopter, gebhardt2018optimizing}. 
A multi-drone approach is presented in \cite{alcantara2020optimal}, obtaining coordinated cinematography trajectories in a distributed way, improving visibility and avoiding occlusions. 

Other solutions adapt classic cinematographic concepts to control the drone in such a way that the recording obtained is visually pleasant. 
Some works focus on finding optimal views considering a static scene, enabling
canonical static shots, like the rule of thirds~\cite{xiong2017automatic}.
Besides the aesthetics, it is often essential to transition between positions in a secure way, keeping a safe distance from the targets~\cite{joubert2016towards}.
%
MPC is used in \cite{nageli2017real} to film scenes while tracking and recording multiple targets, according to some cinematographic standards, i.e., position of targets on the image. This approach is extended to a multi drone context \cite{nageli2017realmulti}, solving  additional challenges like drone-to-drone occlusions and collisions. 
In \cite{alcantara2020autonomous}, a multidrone platform  lets the user choose among a list of canonical drone shots according to \cite{smith2016photographer}.
Recently,~\cite{kratky2021autonomous} presents a control technique for a team of robots while they record a target under various lighting sources. 

 Finally, combining both goals, the solution presented in \cite{bonatti2020autonomous} offers a fully autonomous drone platform that is able to follow an actor while obtaining aesthetic film footage, thanks to avoidance of occlusions, obstacles and jerky trajectories together with fulfillment of canonical shots requirements. 


All these solutions 
focus on getting the best shots by optimizing the extrinsic parameters, e.g., position and orientation of the drones. 
Our approach is the first that takes into account one of the most important factors for high quality photography: the intrinsic parameters of the camera lens.
As detailed next, we extend the existing control problem with this new concept, opening the possibility to control cinematographic trajectories including control of the depth of field, focal distance, and zoom.




 \section{Cinematographic agents}
\label{sec_agents}

This section describes the main agents, placed in the 3D world, inside the CineMPC framework, namely, the drone, the scene, and the camera, and their dynamic models.

\subsection{Drone + gimbal}
The drone and gimbal move and orientate the cinematographic camera in the environment. 
Since our principal aim is to control the intrinsic parameters of the camera, we consider a simplified control model of the pair. 
In particular, we model the state of the drone considering only its position, $\mathbf{p}_{d,k} \in \mathbb{R}^{3},$ and velocity, $\mathbf{v}_{d,k} = \mathbf{\dot{p}}_{d,k} \in \mathbb{R}^{3}$, 
where sub-index $k$ denotes the discrete-time instant $k$.

The gimbal takes care of the camera orientation, $\mathbf{R}_d \in SO(3)$, counteracting the actual drone orientation as well.
All these elements are grouped into the state vector, $\mathbf{x}_{d,k}$,
\begin{equation}
\mathbf{x}_{d,k} = (\mathbf{p}_{d,k}, \mathbf{v}_{d,k}, \mathbf{R}_{d,k}).
\end{equation}
%
The actuators are represented by $\mathbf{u}_{d,k}$, which includes the drone acceleration, $\mathbf{a}_{d,k} \in \mathbb{R}^{3}$, and the gimbal angular velocity, $\mathbf{\Omega}_{d,k}  \in \mathbb{R}^{3}$,
\begin{equation}
\mathbf{u}_{d,k}=(\mathbf{a}_{d,k}, \mathbf{\Omega}_{d,k}).
\end{equation}
Finally, the state transition function, $g$, describes the evolution of the drone state,
\begin{equation}
\mathbf{x}_{d,k+1} = g(\mathbf{x}_{d,k}, \mathbf{u}_{d,k}).
\label{eq:dynamics_drone}
\end{equation}

Note that this simplified model does not hinder real implementation of our solution in a drone. Most current gimbals include stabilization methods that neutralize even the most aggressive drone rotations~\cite{kang2018flycam}. Similarly, there are low-level drone controllers able to follow smoothly high-level trajectories~\cite{mellinger2011minimum}. On the other hand, using simplified models in the control framework enables longer planning horizons and consideration of more complex cost functions.

\subsection{Scene}
The scene is a complicated entity in which a wide variety of complex elements participate, e.g., foreground, background, characters, etc. 
For simplicity, we model the scene as a set of $n$ targets that represent the points of interest to be recorded by the flying camera. 
Similar to the drone, the state, $\mathbf{x}_{t,k}$, of each target $t \in [1,n]$ is described by its position, $\mathbf{p}_{t,k} \in \mathbb{R}^3,$ rotation with respect to the world, $\mathbf{R}_{t,k} \in \mathbb{R}^3,$ velocity, $\mathbf{v}_{t,k} \in \mathbb{R}^{3}$ and angular velocity, $\mathbf{\Omega}_{t,k}  \in \mathbb{R}^{3}$,
\begin{equation}
\mathbf{x}_{t,k} = \left(\mathbf{p}_{t,k} , \mathbf{R}_{t,k},\mathbf{v}_{t,k},\mathbf{\Omega}_{t,k}  \right),
\end{equation}
enabling CineMPC to work with dynamic scenes.

We assume that the targets are visible through the whole recording, leaving the handle of possible occlusions between targets for future work.
Additionally, since the perception of the scene is not a contribution of the work we use ground truth values corrupted with noise in our experiments, noting that this information could be obtained using any
existing perception algorithms, e.g.,~\cite{zakharov2019dpod, seichter2020multi,Kim21ICRA}. 

%


\subsection{Cinematographic camera}
The cinematographic camera is the main component of CineMPC. It is in charge of recording the scene while fulfilling the artistic and technical objectives.

The cameras of most of existing robotic solutions follow the pin-hole model, which only considers projection and geometric parameters. However, the cameras used for cinematography implement the thin-lens camera model. This model includes intrinsic parameters associated with the lens, namely focal length, focus distance, and focus aperture.

The focus distance, $F_k$, represents the space between the camera and the element of the scene where the image is in focus. 
The focal length, $f_k$, measures the distance in millimeters from the camera sensor to the lens of the camera. It affects several artistic features, such as the zoom of the image, the position of the elements of the scene in the image, and the depth of field. The aperture of a lens, $A_k$ indicates how much light reaches the sensor of the camera and is expressed in terms of the f-stop number. The aperture of the lens directly affects the focus of the image and how big the depth of field is.
Therefore, the state of the cinematographic camera of our system is denoted  by $\mathbf{x}_{c,k}$ and includes the aforementioned intrinsic parameters,
\begin{equation}
\mathbf{x}_{c,k} = (f_{k}, F_{k}, A_{k}).
\end{equation}
Large variations of these parameters in a short time lead to abrupt image changes that are not desirable in cinematography. As these parameters can be set to any value as long as they are inside the camera range, instead of acting directly over the intrinsic parameters our framework controls their velocities,
\begin{equation}
\mathbf{u}_{c,k}=(v_{f,k}, v_{F,k}, v_{A,k}),
\end{equation}
where $v_{f,k} \in \mathbb{R}$ expresses the velocity of the focal length, $v_{F,k} \in \mathbb{R}$ represents the velocity of the focus distance and $v_{A,k} \in \mathbb{R}$ is the velocity of the focus aperture, all of them measured in the discrete time step $k$.
Finally, the state transition function, $h$, describes their evolution
\begin{equation}
\mathbf{x}_{c,k+1} = h(\mathbf{x}_{c,k}, \mathbf{u}_{c,k}).
\label{eq:dynamics_camera}
\end{equation}

\section{Control Problem}
\label{sec_control_problem}
With the aim of getting the best image while recording a scene satisfying composition and artistic rules, CineMPC solves a non-linear optimization problem inside an MPC framework. 
MPC is a method to control a process according to a cost function while satisfying a set of constraints. It optimizes a finite time-horizon of $N$ steps of $dt$ seconds each, implementing the current time-slot. Then it optimizes again the $N$ steps taking into account the introduced changes in the system's state \cite{camacho2013model}. In CineMPC, the decision variables are the control inputs described in the previous section whereas the cost function  includes a variety of terms that account for many different camera configurations and artistic effects.

More in detail, at a given time $k_0$, CineMPC solves the following problem over a time horizon $N$,
\begin{equation}
\begin{aligned}
\min_{\substack{\mathbf{u}_{d,k_0}..\mathbf{u}_{d,k_0+N}\\ \mathbf{u}_{c,k_0}..\mathbf{u}_{c,k_0+N}}} \quad & \sum_{k=k_0}^{k_{0}+N}{J_{DoF,k}
    + J_{im,k} + J_{p,k}}\\
\textrm{s.t.} \quad & (\ref{eq:dynamics_drone})\ \hbox{and}\ (\ref{eq:dynamics_camera})\\
  & \mathbf{x}_{d,k} \in \mathcal{X}_d,\ \mathbf{x}_{c,k} \in \mathcal{X}_c\\
  & \mathbf{u}_{d,k} \in \mathcal{U}_d, \mathbf{u}_{c,k} \in \mathcal{U}_c
\end{aligned}.
\end{equation}
The sets $\mathcal{X}_d$ and $\mathcal{X}_c$ represent the constraints over the extrinsic and intrinsic parameters respectively, e.g., avoiding collisions with the targets or maintaining the focal length inside the physically acceptable values of the camera.
Similarly, $\mathcal{U}_d$ and $\mathcal{U}_c$ are used to impose constraints over the control inputs, e.g., feasible drone accelerations or maximum intrinsic velocities to prevent abrupt image changes. 

The cost function is composed by $J_{DoF}$, $J_{im}$, and $J_{p}$. These cost terms, which are associated to the depth of field, the artistic composition and the relative position between  camera and targets, respectively, are described next in more detail.

\subsection{Focus of the image - Depth of Field }  
Our solution autonomously controls the camera depth of field, which represents the space of the scene that appears acceptably in focus in the image. 
According to specialized literature in cinematography and optics~\cite{greenleaf_dof}, ~\cite{bass2010handbook}, the image depth of field is delimited by two points, the near, $D_{n}$ and 
the far, $D_{f}$ distances. 
The space of the scene between $D_{n}$ and $D_{f}$ is in focus, the rest appears blurry in the image.

In order to relate these distances to the camera intrinsics it is convenient to describe first the HyperFocal Distance, $H_{k}$. Placing the focus distance at that distance, the depth of field gets its maximum amplitude. $H_{k}$ is calculated as follows: 
 \begin{equation}
 H_{k} = \frac{f_k^{2}}{A_k c} + f_k,
 \end{equation}
 where $c$ is the circle of confusion, a constant parameter that depends on the model of the camera and expresses the limit of acceptable sharpness, affecting to the field of view.
The near distance, $D_{n,k}$, represents the closest distance to the camera where the focus of the  projected points is acceptable, 
\begin{equation}
D_{n,k} = \frac{F_k (H_{k} - f_k)}{H_{k} + F_k - 2 f_k}.
\end{equation}
Analogously, the far distance, $D_{f,k},$ is the farthest distance to the camera where projected points are acceptably in focus, 
\begin{equation}
D_{f,k} = \frac{F_k (H_{k} - f_k)}{H_{k} - F_k}.
\end{equation}

To determine the desired part of the scene to be in focus, we define the desired near, $D_{n,k}^{*}$, and far distances, $D_{f,k}^{*}$, expressed in meters from the camera.
The cost term of the depth of field in the time step $k$ penalizes intrinsic values that make actual distances depart from the desired values,
\begin{equation}
J_{DoF,k} = w_{D_{n}}\left(D_{n,k} - D_{n,k}^{*}\right)^2 + w_{D_{f}}\left(D_{f,k} - D_{f,k}^{*}\right)^2,
    \label{eq:dof-cost}
\end{equation}
where $w_{D_{n}}$ and $w_{D_{f}}$ are the weights associated to the cost terms of the near and far distances.
It is important to remark that $D_f$ goes to infinite with low values of $f_k$, meaning that the image background is in focus. On the other hand, it is always possible to control $D_n$. This is why $w_{D_{n}}$ is usually higher than $w_{D_{f}}$, or it would be reasonable to give $w_{D_{f}} = 0$ when $f_k$ has a reasonably low value.

\subsection{Artistic composition - Position of elements in the image} 
\label{sec_control_artistic}
The objective of this term is to show the targets placed in particular regions of the image. This term makes the elements appear in the final image so that they satisfy some cinematographic composition rules, e.g., rule of the thirds. 
Using the camera projection model, we define a cost term that penalizes deviations from the desired image composition.
Let $\mathbf{K}$ be the calibration matrix of the camera~\cite{szeliski2010computer},
\begin{equation*}
\mathbf{K}=
\begin{bmatrix}
\beta f_k && s && c_u
\\
0 && \beta f_k && c_v
\\
0 && 0 && 1
\end{bmatrix},      
\end{equation*}
with $c_u$ and $c_v$ the image optical center coordinates and $s$ the skew. 
The focal length is affecting the projection, thus coupling the depth of field and artistic composition objectives.
The parameter $\beta$ is another constant necessary to transform the units of the focal length, given in millimeters, to pixels.
Specifically, the ratio, $\beta = \frac{H_{px}}{H_{mm}} = \frac{W_{px}}{W_{mm}}$, relates the height, $H_{mm}$, and the width, $W_{mm}$, of the camera sensor in millimeters with the height, $H_{px}$, and the width, $W_{px}$, of the image in pixels. 
The projection also requires the relative position between the camera and the target $t$, denoted by $\mathbf{p}_{dt,k}=\mathbf{R}_{d,k}^T\left(\mathbf{p}_{t,k} - \mathbf{p}_{d,k}\right)$. 
The target position in the image, $\textbf{im}_{t,k} \in \mathbb{R}^2,$ is
\begin{equation}
\textbf{im}_{t,k}=
\lambda\mathbf{K}\
\mathbf{p}_{dt,k},
\end{equation}
where $\lambda$ is the normalization factor to remove the scale component in the projection.

Finally, the cost term penalizes deviations with respect to desired image compositions, $\textbf{im}_{t,k}^*,$ for all scene targets $t$,
\begin{equation}
J_{im,k} = \sum_{t=1}^{n}{w_{im,t}\left(\textbf{im}_{t,k} - \textbf{im}_{t,k}^{*}\right)^2},
    \label{eq:im-cost}
\end{equation}
where $w_{im,t}$ is the weight associated to the cost of target $t$.



A target may be defined by several image coordinates, e.g., face, body of a person, etc. Our solution also considers the possibility of controlling the position on the image of different parts of a target, e.g., a person's face matches the upper right third while the knees match the bottom right third. 
According to cinematographic literature, the way in which the scene is recorded strongly affects the sensations that the spectator feels when visualizing it~\cite{thompson2009grammar}. The canonical shots are defined by a set of characteristics with which a shot transmits different feelings to the viewer, e.g., Long shot, Close-Up, Cowboy shot. 
Combined with the position and orientation in which the camera records the target, which is described in the next subsection, placing different parts of the target in the image helps us to record a subject following the features of the canonical shots.

\subsection{Relative position camera-target - Canonical shots} 
To complete a canonical shot feature, we need to record the target from
a certain distance to the camera, or target's depth, and a relative target orientation with respect to it.

The target's depth, $d_{t,k},$ is the only position-related value that cannot be controlled through $J_{im}.$ When combined with a certain value of the focal length, $d_{t,k}$ affects the amount of effective background visible and the image focus level.

The relative rotation between camera and target,
$\mathbf{R}_{dt,k}=\mathbf{R}_{d,k}^T\mathbf{R}_{t,k},$
determines the filming perspective.
In the control problem it is required to enable wide-angle and another aerial type of shots. 
Thus, this cost term is defined in terms of these two parameters and their desired values, $d_{t,k}^{*}$ and $\mathbf{R}_{dt,k}^{*}$,%
\begin{equation}
J_{p,k} = \sum_{t=1}^{n}w_{R}
\left\|\mathbf{R}_{dt,k}^T - \mathbf{R}_{dt,k}^{*}\right\|_F  + w_{d}\left(d_{t,k} - d_{t,k}^{*}\right)^2,
\label{eq:kind-cost}
\end{equation}
with $w_{R}$ and $w_{d}$ the corresponding weights and $\left\|x\right\|_F$ calculates the Frobenius norm of $x$.



\section{Experimental Validation}
\label{sec_experiments}
To show the potential of our solution, we present a complete cinematographic recording composed by seven parts. Each part 
demonstrates different artistic requirements or specifications that CineMPC helps to fulfill.
Next we describe the implementation and execution details and discuss the experiment content and analysis.

\subsection{Implementation details}
The system is implemented in C++ and communicates with CinemAirSim \cite{pueyo2020cinemairsim}. Thus, this work is validated in a photorealistic scenario where drones have realistic physics and dynamics. AirSim implements a low-level controller for the drone that takes a trajectory of 3D world points and a desired velocity as input and moves the drone accordingly. CineMPC calculates a set of 3D points and velocities according to the actuators that this low-level controller transforms into a feasible trajectory and gives the commands to the drone.
Since the proposed optimization problem is non-linear, we use Ipopt (Interior Point OPTimizer) \cite{wachter2006implementation} to solve the MPC problem.
The ground-truth data used for the experiments is obtained from AirSim. In order to make experiments closer to reality, i.e. to simulate more realistic measures, the provided position of the targets is perturbed with a Gaussian noise of zero mean and standard deviation of $\sigma = 0.04m^2,$ for each component. Based on experimentation and knowledge, a Gaussian noise with these values introduces a reasonable perturbation that the system can absorb without lowering its performance.  The orientation of the targets is perturbed with a Gaussian noise of zero mean and standard deviation of $\sigma = 0.01$rad$^2$.  These noisy measures are filtered with a Kalman Filter in order to reduce the noise impact in the final result, as we would do in real world scenarios. The filter would also be useful for robust estimation of the pose of the targets in case of a miss-detection in a real case scenario. 

The system is composed of different modules. One module is in charge of reading the state of the system, another module runs the MPC according to this state and user's instructions and calculates the next $N$ discrete steps, a third module joints these steps creating a more continuous and softer spline trajectory that gives to the drone. Each module is run in a different thread, the main thread handles the communication and synchronization between them. 

All the experimentation is run in an Intel® Core™ i7-9700 8-Core CPU equipped with 64 Gb of RAM and a NVidia GeForce GTX 1070.
We use a sample period of $dt = 0.3 s.$ and time-horizon of $N = 5$ time-steps.
Table \ref{table:camera-params} contains the set of constants described in Sec. \ref{sec_agents} that the camera of the simulation environment implements.
\begin{table}[!hb]
\begin{center}
\caption{Camera Parameters}
\begin{tabular}{| c | c | c | c | c | c | c | c | c | }
\hline
\multicolumn{2}{|c|}{$px$} & \multicolumn{2}{|c|}{$mm$} & $\beta$ & $c_u$ & $c_v$& $s$& $c$
\\
\hline 

W & H & W & H & \multirow{ 2}{*}{0.5625}    &\multirow{ 2}{*}{480}   & \multirow{ 2}{*}{270}  &\multirow{ 2}{*}{0} &\multirow{ 2}{*}{0.03}

 \\
 \cline{0-3}

960 & 540 & 13.365 & 23.76 &  &  &  & & \\ 
\hline
\end{tabular}
\label{table:camera-params}
\end{center}
\end{table}

\subsection{Complete cinematographic example}

We define the desired values of each experiment following cinematographic literature~\cite{thompson2009grammar},~\cite{brown2016cinematography} to achieve artistic and technical guidelines to show how CineMPC easily records footage with different requirements.  

This footage records two people in a park, a man and a woman, targets of the scene. To standardize the notation, the state of the targets, are represented by $\mathbf{x}_{m}$ and $\mathbf{x}_{w}$. The man position and rotation are described as $\mathbf{p}_m \in \mathbb{R}^{3}$ and $\mathbf{R}_m \in \mathbb{R}^{3}$ converted to roll, pitch, yaw notation, respectively. From $\mathbf{p}_m$ we estimate the position of different parts of his body (nose($n\_m$), knees($l\_m$), etc) so we can place the target in the image to meet the requirements of the canonical shots as explained in Sec. \ref{sec_control_artistic}. The man moves trough the scene with a linear velocity of $\mathbf{v}_{m,k}\in \mathbb{R}^{3}$ and rotates at some point with an angular velocity of $\mathbf{\Omega}_{m,k} \in \mathbb{R}^{3}$.
Thus, the state of the man in the time-step $k$ is defined by 
$\mathbf{x}_{m,k} = \left(\mathbf{p}_{m,k} , \mathbf{R}_{m,k},\mathbf{v}_{m,k},\mathbf{\Omega}_{m,k}\right).$
Analogously, the state of the woman is represented by $\mathbf{x}_{w,k} = \left(\mathbf{p}_{w,k} , \mathbf{R}_{w,k},\mathbf{v}_{w,k},\mathbf{\Omega}_{w,k}\right).$ 
The goal is to keep the drone always close to the targets, so the desired relative distance to the tracked target at each time is $d_{t,k}=5m.$, and its weight is set to $w_{t,k}=1$ in the whole experiment. 
As described in Sec. \ref{sec_control_problem}, we set different weights to the cost terms to normalize their measurement units and to allow to configure higher relevance (more weight) to some cost terms or targets. First row of table \ref{table:weights} shows the weights used to standardize the units that are common to all the sequences. The rest of the rows show the weights for each of the analyzed sequences depending on the importance of that term in that particular footage. Some weights can be zero if we do not want to control that cost term in that sequence. 

As baseline for comparison, we run the same experiment without the camera intrinsics in the control problem, but setting their constraints to a fixed value, so CineMPC cannot optimize their value, e.g., $f_k$ = [35, 35] mm. in $\mathcal{X}_c$.

The full footage is composed by seven sequences, which can be considered as seven different experiments, as the requirements and desired values vary from one to another. We refer the reader to the supplementary video to watch the full recording\footnote{\url{https://youtu.be/1kV-5nLl6BI}}.
Next, we describe and discuss only the three most representative sequences (second, third, and fifth)
according to their technical and artistic complexity. Each sequence description is accompanied with different figures to show its qualitative and quantitative results and analysis. We recommend to visualize these figures with zoom for better appreciation of the details.
\begin{table}[t]
\begin{center}
\caption{Weights of cost function terms}
\begin{tabular}{| c | c | c | c | c | c | c | }
\hline
   & $w_{D_{n}}$ & $w_{D_{f}}$ & ${w_{im,m}}$ & ${w_{im,w}}$ & ${w_{R,m}}$& ${w_{R,w}}$ \\ \hline
 units & 10 & 10 & 1 & 1 & 5000 & 5000 \\
 \hline
 seq.2 & 10 & 0 &  1  & 0  & 10000 & 0 \\
seq.3 & 50 & 50 & 1 & 0 & 5000 & 0  \\
seq.5 & 10 & 10 & 2 & 2 & 2500 & 2500 \\ \hline
\end{tabular}
\label{table:weights}
\end{center}
\end{table}

\subsubsection{\textbf{Depth of field control}. Third sequence}
This experiment highlights how CineMPC successfully controls the depth of field with the camera intrinsics. The goal is to only keep the man in focus, blurring the rest of the scene.
This effect is achieved by drastically narrowing the depth of field to the part of the scene where the man is placed. In particular, the desired near and far distances are set to 2.5 meters behind the man and one meter in front of him, $D_{n,k}^{*} = \lVert\mathbf{p}_{m,k}\rVert -2.5m$ and $D_{f,k}^{*} = \lVert\mathbf{p}_{m,k}\rVert + 1m.$ The weight $w_{D_f}$ is higher to emphasize the desired effect.

\begin{figure}[!t]
\centering
\begin{tabular}{cc}
    \includegraphics[width=0.46\columnwidth,height=2.4cm]{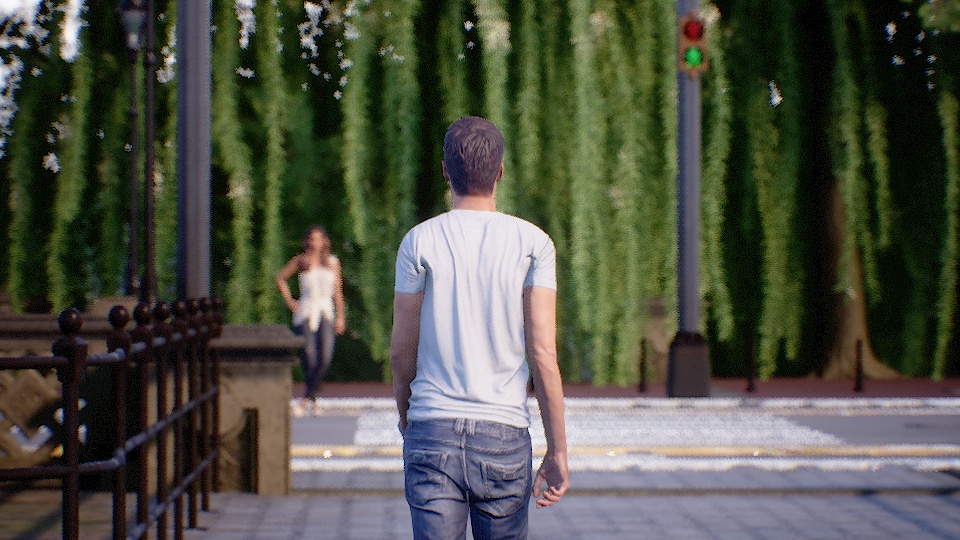}
    &
    \includegraphics[width=0.48\columnwidth,height=2.6cm]{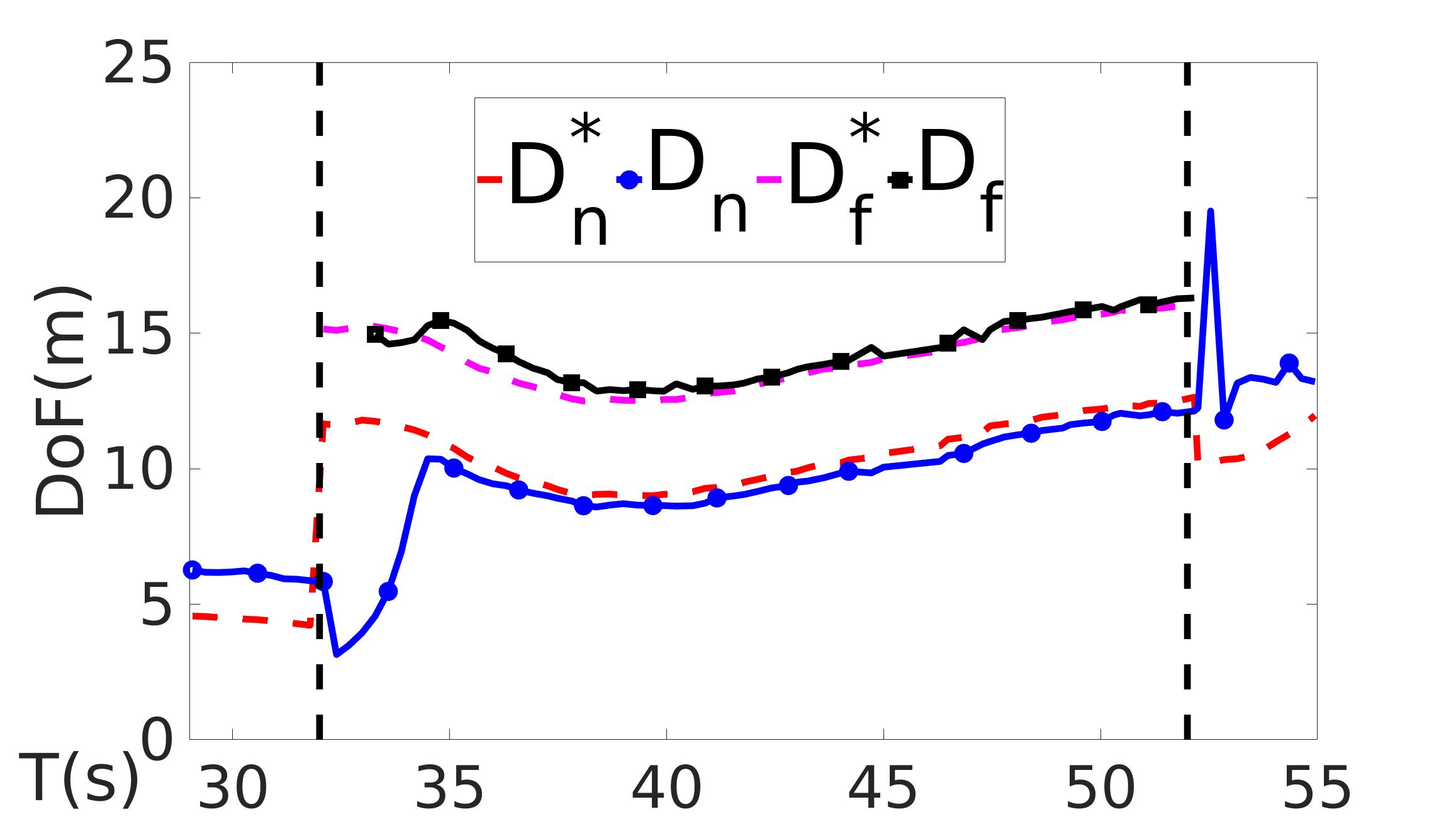}
   \\
    \footnotesize (a) CineMPC & \footnotesize (c) $D_n$ and $D_f$
    \\
       \includegraphics[width=0.46\columnwidth,height=2.4cm]{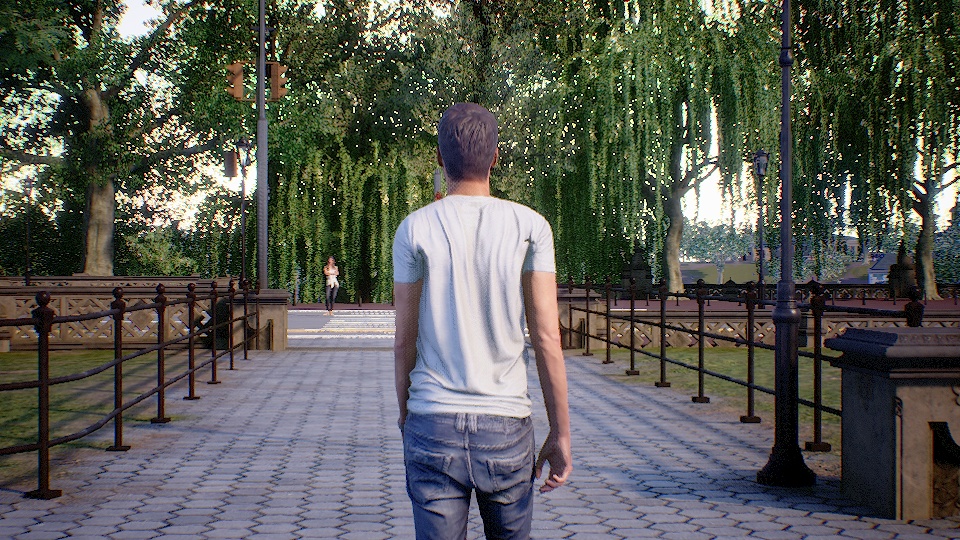}
        
    &
  \includegraphics[width=0.50\columnwidth,height=2.6cm]{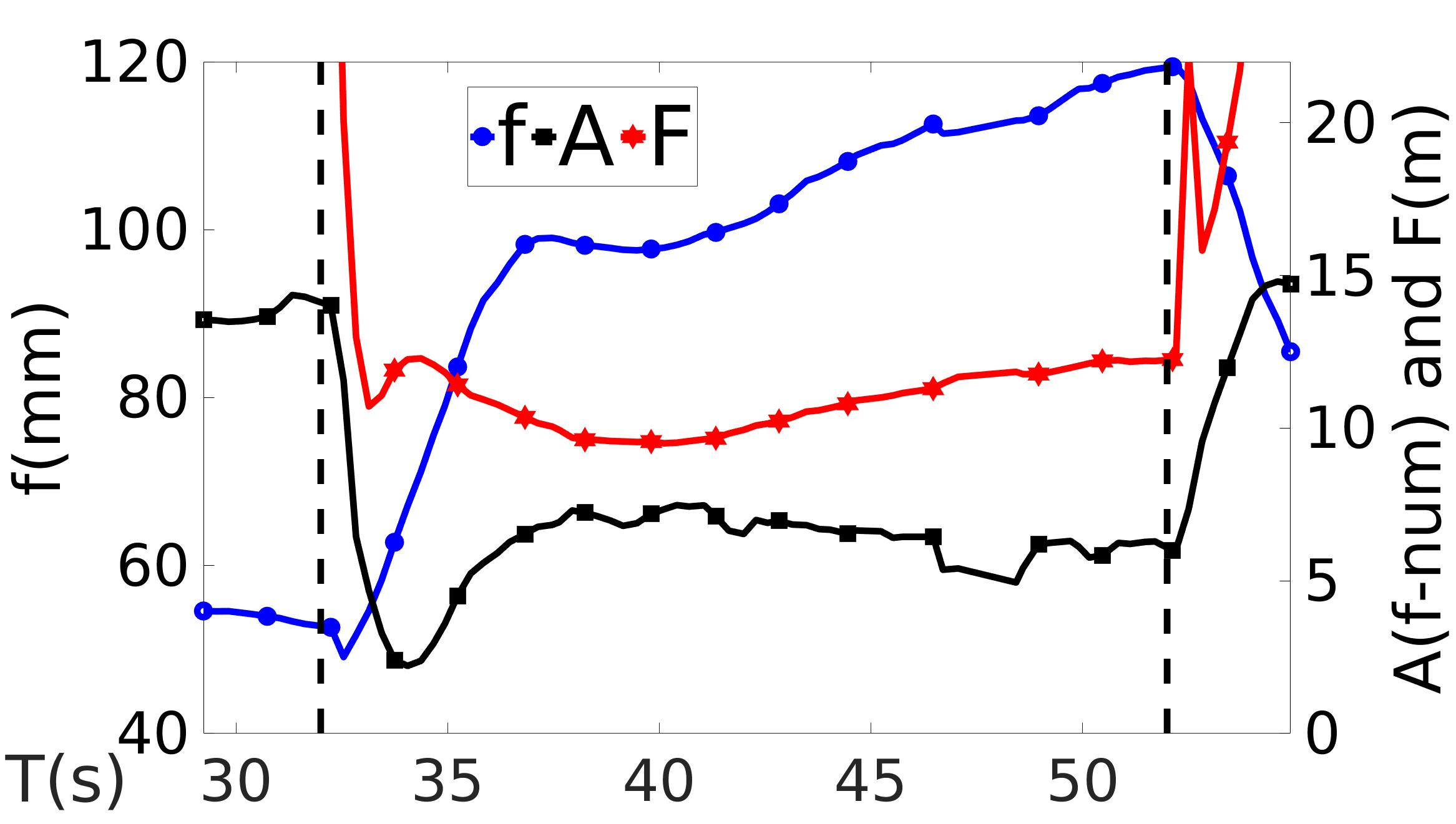}\\
     \footnotesize (b) Baseline & \footnotesize (d) Camera intrinsics
\end{tabular}
\caption{\textbf{Depth of field control.} (a)  Sequence frame obtained with CineMPC. (b) Same frame without optimizing the intrinsics. (c) Evolution of the depth of field ($D_n$ and $D_f$) values. Solid 
lines represent actual values and dashed lines desired values. (d) Values of the camera intrinsics along the sequence. In both plots dashed vertical lines denote the sequence start and end. 
}
\label{fig:seq3}
\end{figure}

Figure~\ref{fig:seq3} shows qualitative results of the experiment using CineMPC (a) and the baseline (b), in particular the image captured at certain time step with each alternative. In (a) the man is the only part in focus, as intended. In (b) the camera intrinsics are not considered and the result is very different.

A quantitative analysis is also shown in Fig. \ref{fig:seq3} (c) and (d). The plot in (c) illustrates the controller ability to track the desired values of $D_n$ and $D_f$, even as the desired values change because of the drone and target motion. 
The achieved effect is explained by the evolution of the intrinsics, displayed in (d). The focal length increases its value while the aperture goes to its minimum value. This increases the hyperfocal distance, achieving the desired narrow depth-of-field.
At the same time, the focus distance is driven to the distance to the target to put the focus on the man, following the same evolution than $D_n$ and $D_f$.

\subsubsection{\textbf{Artistic composition.} Fifth sequence}
This sequence demonstrates how to obtain different image compositions through $J_{im}$, controlling the position of the targets in the image.
In this example we control the position of two vertical points of the woman with the aim of recording her using a \textit{'Cowboy Shot'}, i.e., place the upper part of her body above the lower third of the image.
We want her whole body centered in the right horizontal third, and her nose and knees, placed in the upper and lower thirds of the image, respectively, $\textbf{im}_{n\_w,k}^{*} = [\frac{2W_{px}}{3}, \frac{H_{px}}{3}]$ and $\textbf{im}_{l\_w,k}^{*} = [\frac{2W_{px}}{3}, \frac{2H_{px}}{3}]$.
We also control the position of the man, placed in the intersection of the left horizontal and the vertical superior thirds, $\textbf{im}_{n\_b,k}^{*} = [\frac{W_{px}}{3}, \frac{H_{px}}{3}]$.

\begin{figure}[!h]
\centering
\begin{tabular}{cc}
    \includegraphics[width=0.43\columnwidth,height=2.2cm]{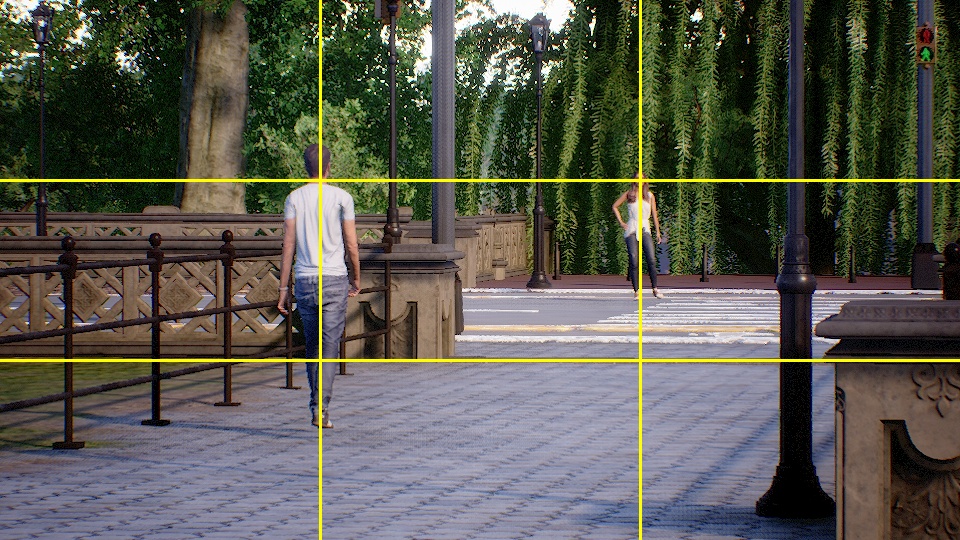}
    & 
    \includegraphics[width=0.52\columnwidth,height=2.4cm]{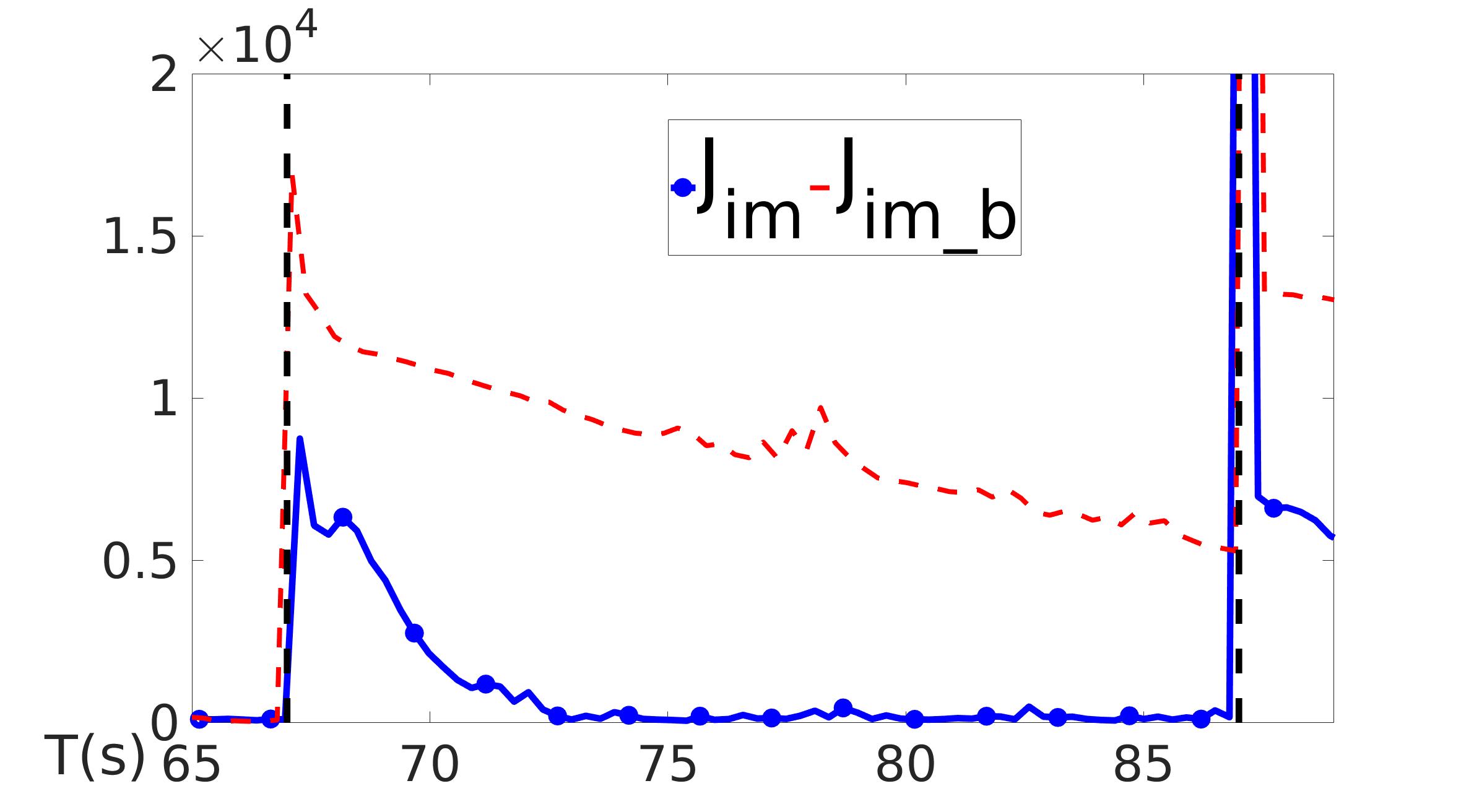} 
     \\\footnotesize (a) Initial sequence frame & \footnotesize (d) $J_{im}$ and $J_{im\_baseline}$
    \\ 
    \includegraphics[width=0.43\columnwidth,height=2.2cm]{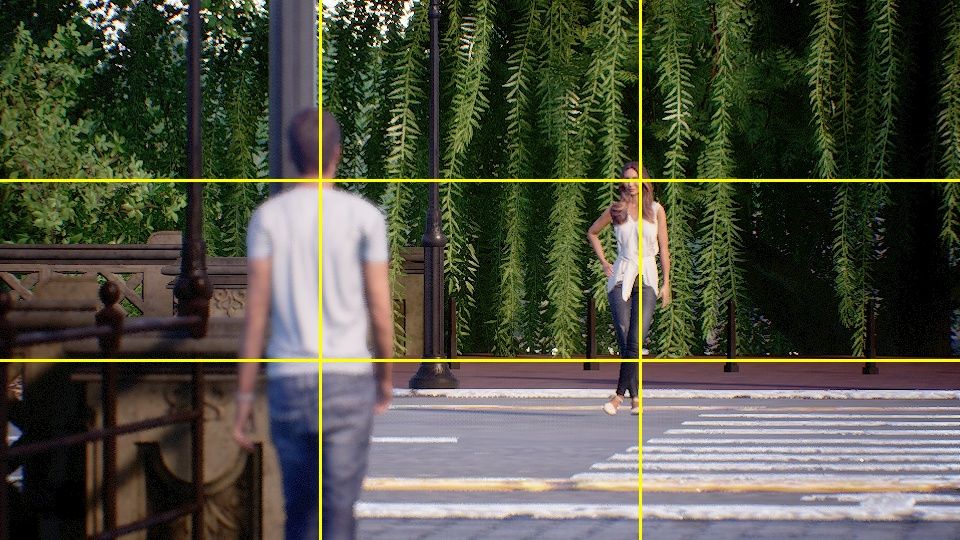}
    &
    \includegraphics[width=0.52\columnwidth,height=2.4cm]{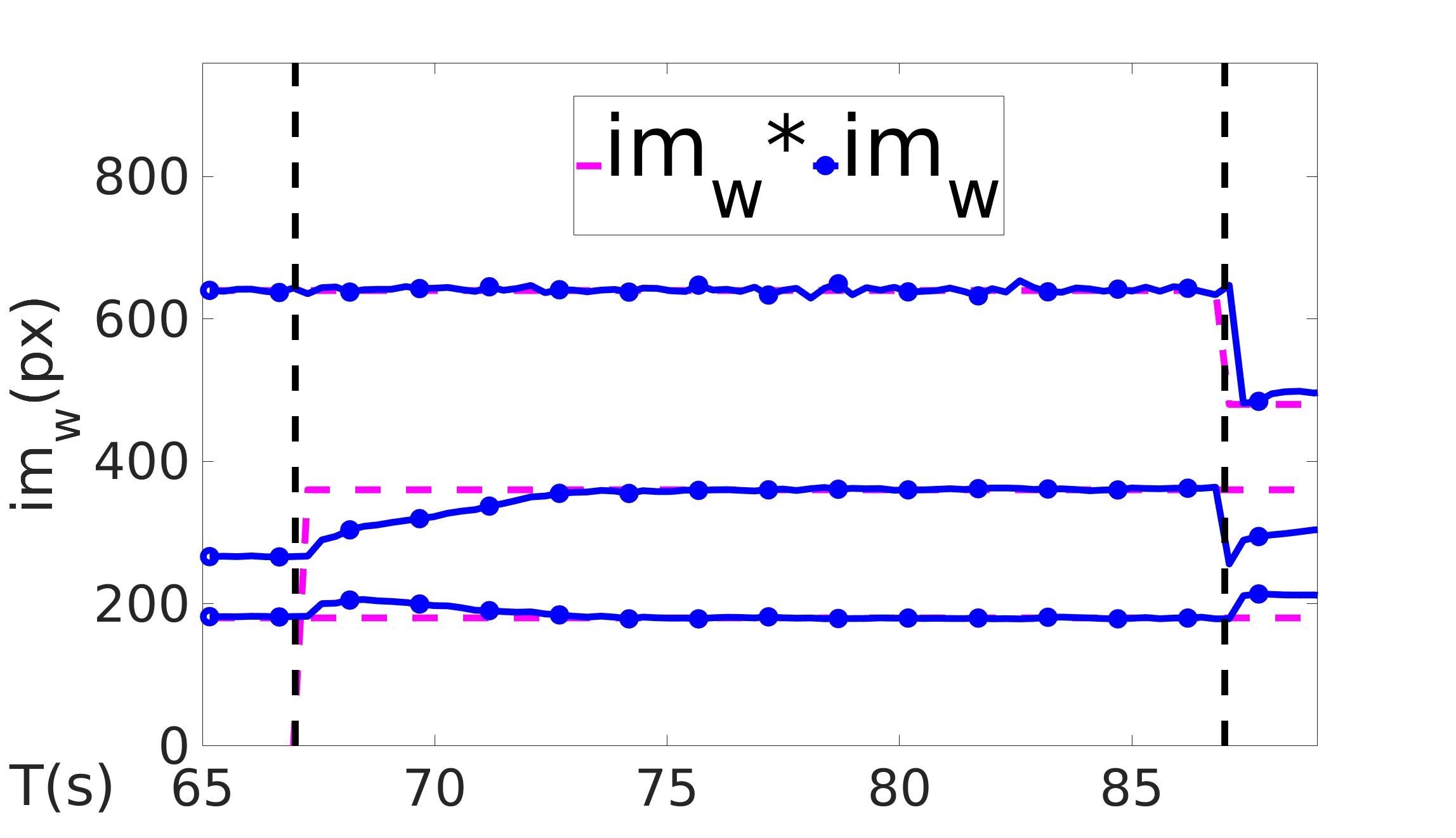}

    \\\footnotesize (b) Sequence frame. CineMPC & \footnotesize (e) $\mathbf{im_w}$\\
    \includegraphics[width=0.43\columnwidth,height=2.2cm]{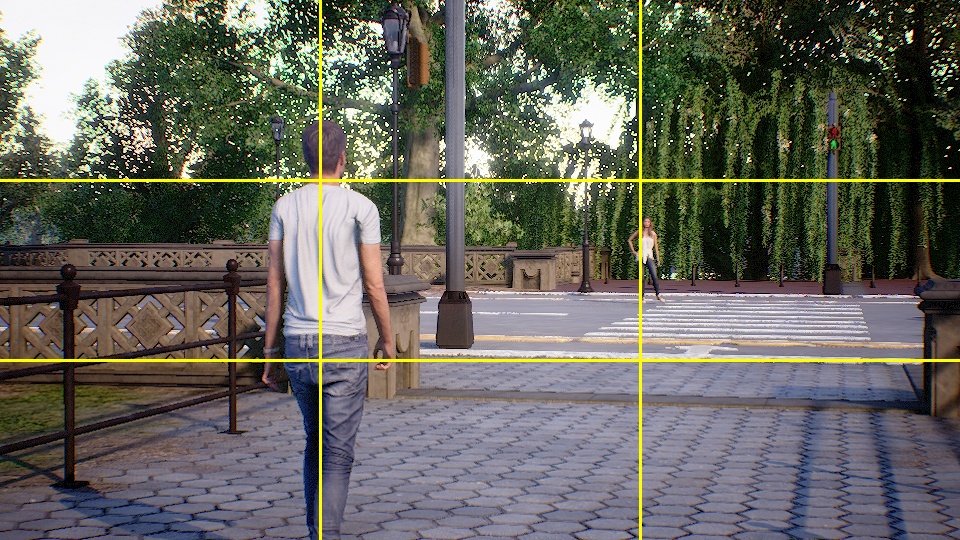}
     &
    \includegraphics[width=0.52\columnwidth,height=2.4cm]{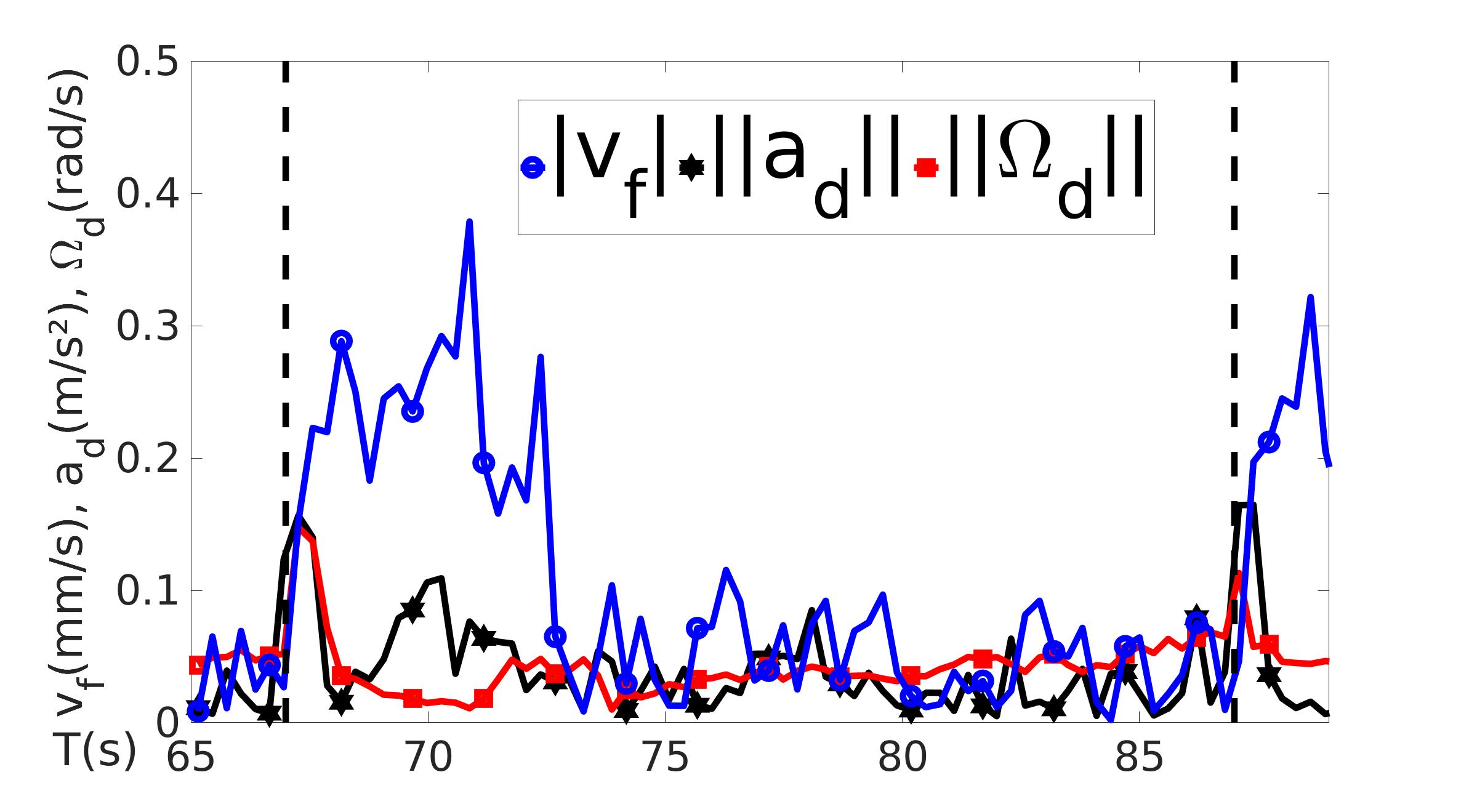}
    
   \\  \footnotesize (c) Sequence frame. Baseline & \footnotesize (f) Actuators
\end{tabular}
\caption{\textbf{Artistic composition.} 
Qualitative results: (a)  Initial sequence frame. (b) Sequence frame where constraints of $J_{im}$ are satisfied. (c) Same frame obtained with the baseline, no intrinsics optimization. Quantitative results: (d) Evolution of $J_{im}$ and baseline value. (e) Evolution of $\mathbf{im_w}$ and desired value $\mathbf{im_w^*}$. Solid lines are actual values and dashed lines are desired values. Top line is horizontal pixel and bottom two lines are vertical pixels (f) Evolution of the actuators that affect $J_{im}$. Dashed vertical lines denote the sequence start and end in the plots.
}
\label{fig:seq5}
\end{figure}

Figure \ref{fig:seq5}(a-c) shows qualitative results of this experiment. The guidelines of the rule of thirds are displayed in yellow in these frames. The initial frame and a frame when the desired values are achieved of this sequence are shown in (a) and (b) respectively, with the body parts of the man and the woman matching the desired thirds. (c) shows the same frame of (b) when the intrinsics are not in the loop of control. 
On the other hand, as seen in the baseline (c) cannot acquire this shot without the focal length, as the drone is physically too far and zoom is needed.

Quantitative results are depicted in Fig. \ref{fig:seq5} (d-f). The first plot (d) shows the evolution of $J_{im,k}$ using CineMPC and the baseline.
It can be seen how this cost cannot be decreased when the focal length is not controlled, the only way of achieving the effect is by moving the drone close to the targets. Fig. \ref{fig:seq5} (e) shows the evolution of image coordinates of the woman, $\mathbf{im}_{w,k}$ 
and their desired values, $\mathbf{im}_{w,k}^*$. Upper line denotes the horizontal pixel ($im.x_{w,k}$) and lower lines represent the vertical pixels of her nose and legs, respectively ($im.y_{n,k}$, $im.y_{l,k}$).
The actuators that make this possible are depicted in Fig. \ref{fig:seq5} (f), namely the normalized drone linear accelerations and angular velocities ($\mathbf{a}_{d,k},  \mathbf{\Omega}_{d,k}$) and the focal length velocity ($\mathbf{v}_{f,k}$). The high focal velocity at the beginning of the sequence increases the focal length and helps to achieve the desired values as shown in Fig. \ref{fig:seq5} (e).

 \begin{figure}[!ht]
\centering
\begin{tabular}{cc}
    \includegraphics[width=0.44\columnwidth,height=2.2cm]{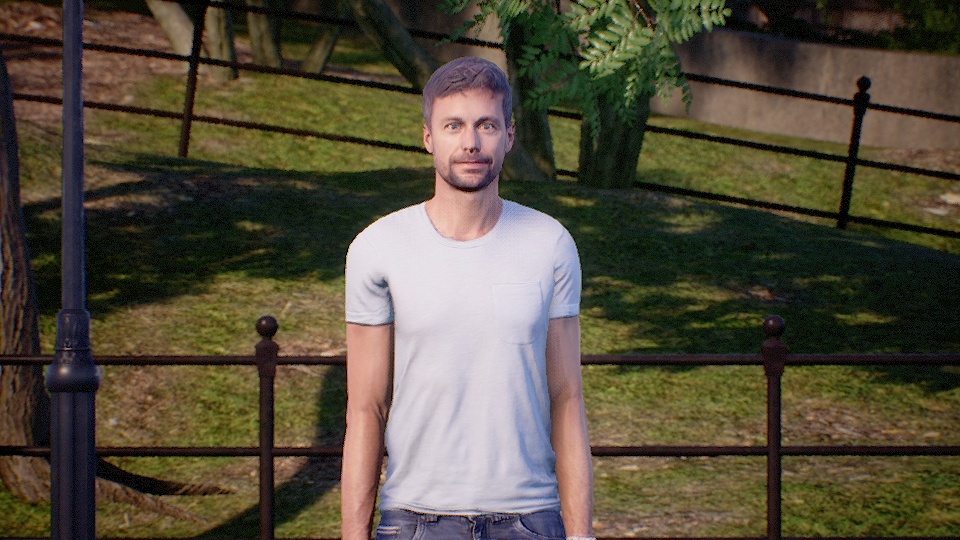} &
    \includegraphics[width=0.44\columnwidth,height=2.4cm]{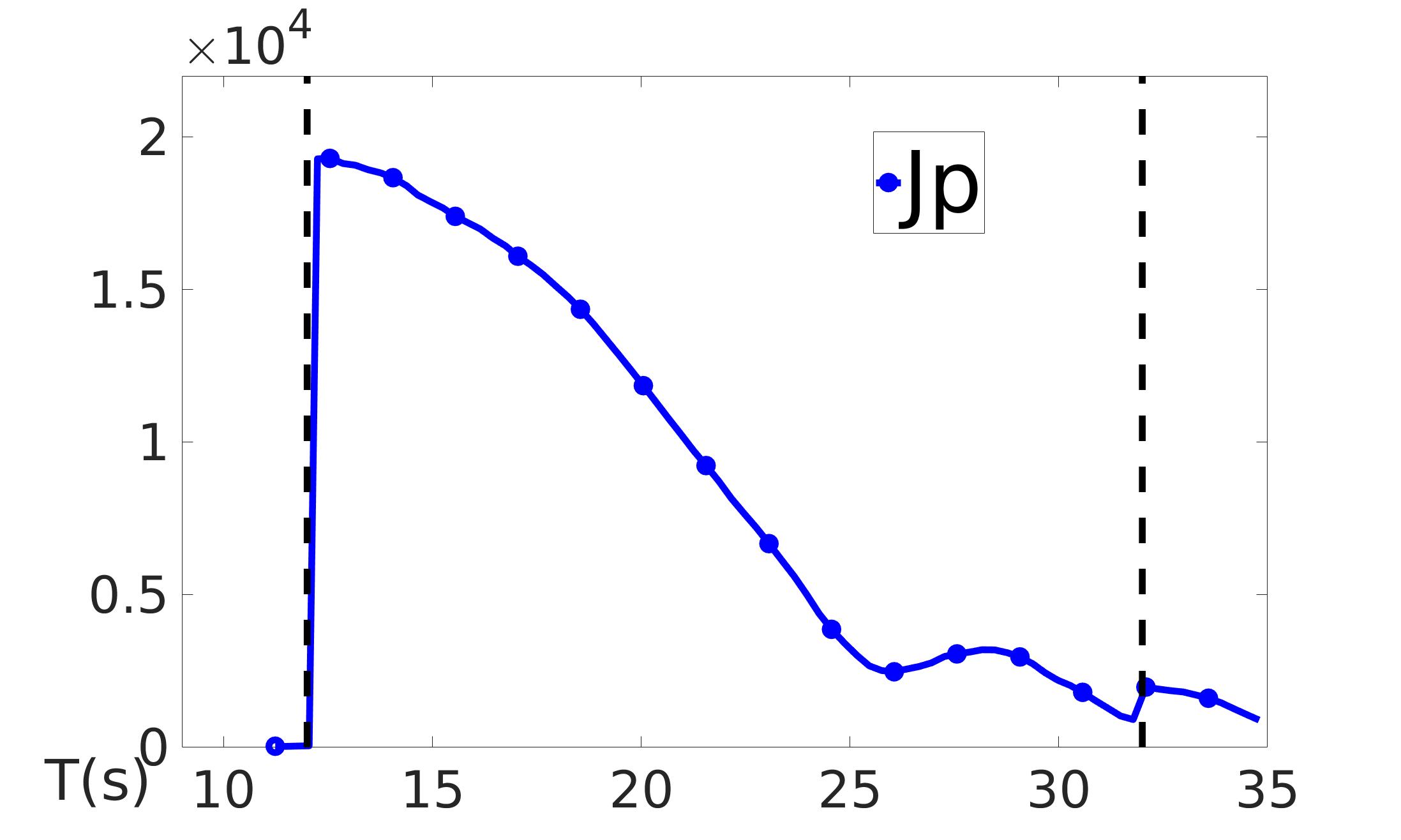}\\
    \footnotesize (a) Initial sequence frame & \footnotesize (c) $J_{p}$ 
    \\
    \includegraphics[width=0.44\columnwidth,height=2.2cm]{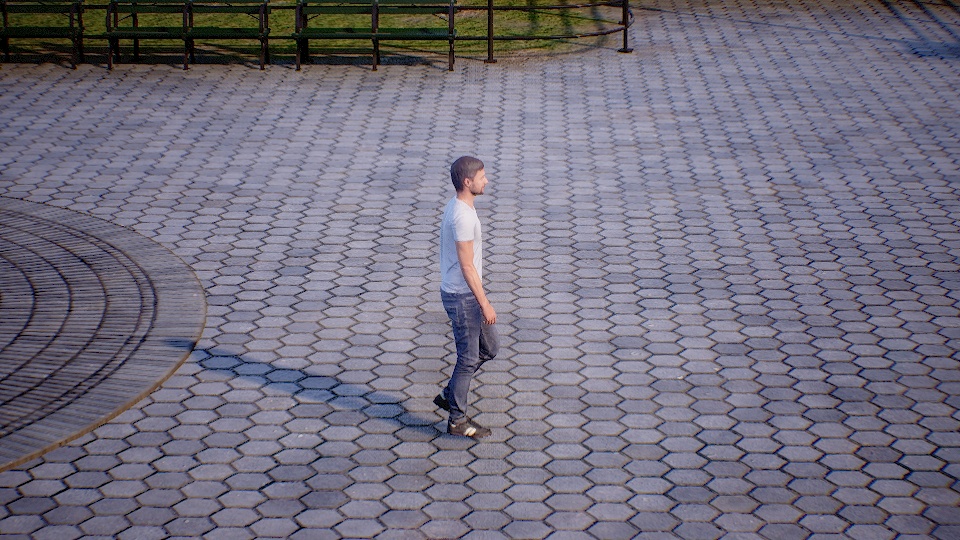} &
    \includegraphics[width=0.44\columnwidth,height=2.4cm]{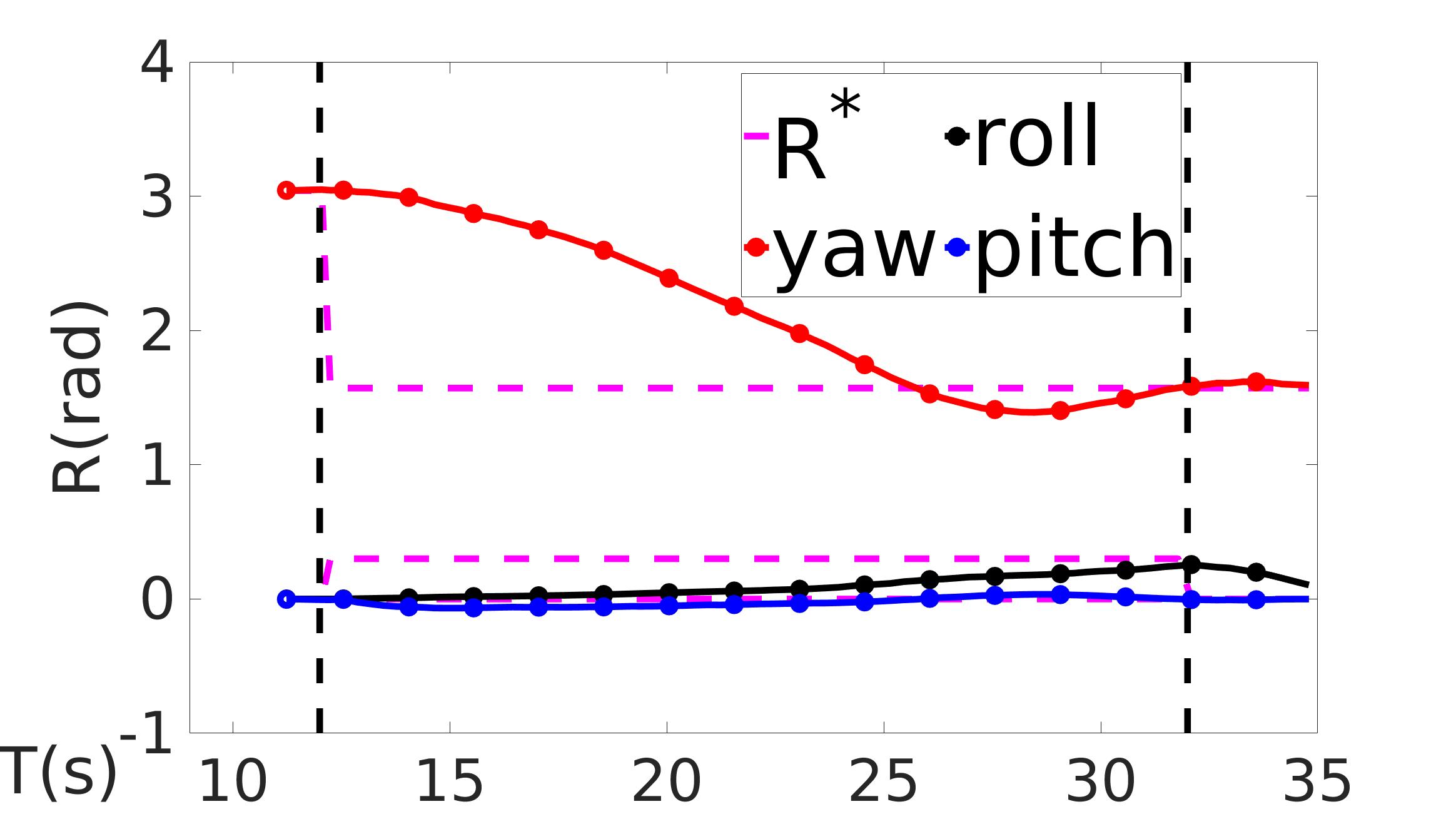}\\
    \footnotesize (b) End sequence frame& \footnotesize (d) $\mathbf{R_{dm}}$ \\
\end{tabular}
\caption{
\textbf{Controlling relative position}. Qualitative results: (a)  Initial sequence frame. (b) End sequence frame. Quantitative results (dashed vertical lines denote the sequence start and end): (c) value of $J_{p}$ along the sequence; (d) values of the relative rotation between the camera and the target along the sequence (solid lines), together with their desired values (dashed lines). 
}
\label{fig:seq2}
\end{figure}

\subsubsection{\textbf{Controlling relative position}. Second sequence}
To illustrate the effect of $J_p$ in the recording, we capture what literature calls a \textit{'High-Angle Shot'}, i.e., record target from up, and from the right. This generates an aerial side view of the man. With this purpose we set the desired relative rotation $\mathbf{R}_{dm,k}^{*} = [0,R_{m.pitch}-0.3, R_{m.yaw}- \frac{\pi}{2}]^T\ R_{m}$ rads in roll, pitch, yaw notation. Although we give less weight to $w_{p,k}$, we set its desired value to $\mathbf{p}_{dm,k}^{*} = 5\ m.$ 
Fig. \ref{fig:seq2}(a,b) shows the start and end frames of the footage, that show how the drone starts looking at the man from the front and same height, and finishes the recording looking at him from his right and up (best seen in supplementary material video).
Fig. \ref{fig:seq2}(c,d) shows a quantitative analysis of this experiment. The first plot (c) shows the value of $J_{p,k}$ during the execution of the sequence, clearly decreasing.
The other plot shows the evolution of the rotations, where the values of roll, pitch and yaw go to their desired values. The rotations take long to get to their desired values due to the impact of the rest of the cost terms (keeping the target centered all the time). The desired value of relative distance is not totally satisfied because the weights of $J_{im}$ and rotations are higher.

\section{Conclusions}
\label{sec_conclusions}

We have presented CineMPC, a model predictive control approach to control the intrinsic and extrinsic parameters of a camera for autonomous cinematography. This is the first approach to date to include the intrinsic information in this kind of control.  
We have described the main cinematographic agents and discussed in detail the control problem. 
It includes three different cost terms to achieve several artistic guidelines, depth of field, artistic composition of the image and canonical shots.
The optimization of these terms returns camera control values that generate semantically expressive images, closer to the ones seen in actual movies.
A complete scene has been used to illustrate the potential of CineMPC in photorealistic simulation, successfully considering time-varying guidelines, perturbing the simulator perception data with noise to make it closer to reality.
Future work will consider implementing CineMPC in a real drone, including a more comprehensive integration with real perception techniques, obstacles and occlusions avoidance.



\balance
\bibliographystyle{IEEEtran}
\bibliography{references}

\end{document}